\documentclass[preprint,12pt]{elsarticle}
\usepackage[a4paper, total={6.5in, 8in}]{geometry}

\usepackage{amssymb}
\usepackage{amsmath}
\usepackage{amsfonts}
\usepackage{mathrsfs}
\usepackage{booktabs}
\usepackage{makecell}
\usepackage[mathscr]{euscript}
\usepackage{hyperref}
\usepackage{lineno}

\begin{document}

\begin{frontmatter}



\title{On the Role of Consistency Between Physics and Data in Physics-Informed Neural Networks} 


\author[label1]{Nicolás Becerra-Zuniga\corref{cor1}}
\ead{n.becerra@upm.es}
\cortext[cor1]{Corresponding author.} 
\author[label2]{Lucas Lacasa}
\author[label1,label3]{Eusebio Valero}
\author[label1,label3]{Gonzalo Rubio\corref{cor1}}
\ead{g.rubio@upm.es}

\affiliation[label1]{organization={ETSIAE-UPM-School of Aeronautics, Universidad Politécnica de Madrid},
            addressline={Plaza Cardenal Cisneros 3}, 
            city={Madrid},
            postcode={28040}, 
            state={Madrid},
            country={Spain}}
\affiliation[label2]{organization={Institute for Cross-Disciplinary Physics and Complex Systems (IFISC, CSIC-UIB)},
            addressline={Campus Universitat de les Illes Balears, Edificio Científico-Técnico}, 
            city={Palma de Mallorca},
            postcode={07122}, 
            state={Islas Baleares},
            country={Spain}}

\affiliation[label3]{organization={Center for Computational Simulation, Universidad Politécnica de Madrid},
            addressline={Campus de Montegancedo}, 
            city={Boadilla del Monte},
            postcode={28660}, 
            state={Madrid},
            country={Spain}}
\begin{abstract}

Physics-informed neural networks (PINNs) have gained significant attention as a surrogate modeling strategy for partial differential equations (PDEs), particularly in regimes where labeled data are scarce and physical constraints can be leveraged to regularize the learning process. In practice, however, PINNs are frequently trained using experimental or numerical data that are not fully consistent with the governing equations due to measurement noise, discretization errors, or modeling assumptions. The implications of such data--to--PDE inconsistencies on the accuracy and convergence of PINNs remain insufficiently understood.
In this work, we systematically analyze how data inconsistency fundamentally limits the attainable accuracy of PINNs. We introduce the concept of a \emph{consistency barrier}, defined as an intrinsic lower bound on the error that arises from mismatches between the fidelity of the data and the exact enforcement of the PDE residual. To isolate and quantify this effect, we consider the one-dimensional viscous Burgers equation with a manufactured analytical solution, which enables full control over data fidelity and residual errors. PINNs are trained using datasets of progressively increasing numerical accuracy, as well as perfectly consistent analytical data.
Results show that while the inclusion of the PDE residual allows PINNs to partially mitigate low-fidelity data and recover the dominant physical structure, the training process ultimately saturates at an error level dictated by the data inconsistency. When high-fidelity numerical data are employed, the PINN solutions become indistinguishable from those trained on analytical data, indicating that the consistency barrier is effectively removed. These findings clarify the interplay between data quality and physics enforcement in PINNs and provide practical guidance for the construction and interpretation of physics-informed surrogate models.
\end{abstract}




\end{frontmatter}


\section{Introduction}

Within the recent advent of machine learning techniques in the physical sciences \cite{carleo2019machine}, physics-informed neural networks (PINNs) \cite{karniadakis2021physics, cai2021physics, zhao2024comprehensive} have emerged as a popular supervised learning framework \cite{goodfellow2016deep}, thanks to its capacity for building data-driven surrogate solvers of partial differential equations (PDEs) under data scarcity. Originally introduced by Lagaris et al. \cite{lagaris1998artificial} and later extended to deep neural networks \cite{goodfellow2016deep} by Karniadakis et al. \cite{raissi2019physics}, the key idea is to incorporate the governing laws --i.e. the extent to which the approximate solution offered by the neural network verifies the PDE-- directly as a term in the loss function of a (deep) neural network. Accordingly, the resulting optimization of PINN parameters is driven not only by the standard learning-from-data principle, but also by the actual PDE the surrogate aims to approximate. Incorporating such a `PDE term' in the loss plays the role of a strong inductive bias and effectively regularises the PINN optimization towards regions of its parameter space that, in turn, might provide enhanced generalization performance: the PINN is effectively forced to ``learn the physics'' and thus is potentially able not only to interpolate well, but also to extrapolate to configurations not included in the training data \cite{bonfanti2024generalization}.
Accordingly, this hybrid approach is particularly relevant in scientific and engineering problems where training data is scarce --e.g. experimental measurements or numerical simulations might be limited, expensive to obtain, or contaminated by noise--, yet the governing equations are well established. Applications range from fluid dynamics \cite{zhao2024comprehensive}, heat transfer \cite{cai2021physics2} or power systems \cite{huang2022applications} to material science \cite{zhang2022analyses}, physiology \cite{sahli2020physics} or electromagnetism \cite{khan2022physics}, to cite some \cite{cuomo2022scientific}.\\
Three representative scenarios illustrate the relevance of PINNs in such contexts. The first corresponds to building surrogate solvers of physical processes from experimental data. While experimental data might sample well the temporal dimension, these are often sparse in space and physical parameters, and are usually affected by measurement noise. The objective of the PINN surrogate is to build a solver that predicts a continuous-time, spatio-temporal field which is consistent both with the available data and the known underlying physics. The second scenario arises in computational modeling --for instance in computational fluid dynamics (CFD) \cite{ferziger2019computational}, where typically we aim to integrate the Navier-Stokes (NS) equations over some fixed geometry and for a concrete instantiation of physical parameters (e.g. Reynolds, Mach, etc)--. Here the task is to build a surrogate solver trained from CFD simulation data. Now, since each numerical simulation is itself computationally expensive, this often precludes the access to CFD data that densely covers the range of different geometries and physical parameters. In these cases, a PINN can be trained on a limited dataset of high-fidelity samples, and its training is regularized by enforcing the relation between physical variables in the PINN to follow NS or some approximate model \cite{sun2020surrogate, eivazi2022physics}. Once successfully trained, the PINN serves as a fast surrogate solver, enabling near-instantaneous predictions at inference across a wide range of conditions. Finally, a third scenario combines both experimental and numerical simulations data in order to build a (larger) multi-fidelity training set over which the PINN can be trained \cite{penwarden2022multifidelity}.

\medskip \noindent
Despite their promise and success, PINNs often face the subtle yet critical challenge of assuring a \emph{consistency} between the physics (i.e. the actual PDE it aims to solve) and the data sources it uses for training. In most formulations, throughout optimization the PINN parameters are updated by leveraging automatic differentiation \cite{raissi2019physics} to compute the gradients of {\it both} loss terms (related to the data residual and the PDE residual respectively). In other words, we are implicitly assuming that the available data --both the one used for training, and the one used to test the performance of the PINN in unseen data-- follows exactly the governing PDE. But neither experimental data nor numerical simulations exactly capture the underlying physical reality. As a matter of fact, available data often originate from heterogeneous sources with uncertainty, whose compounded effect make them potentially deviate from the above mentioned assumption. For instance, experimental data, while in principle consistent with the PDE, suffers from several types of measurement errors. Likewise, numerical simulations data, while free from measurement errors, can be affected by algorithmic issues (from discretization errors to biases induced by the specific algorithmic choices for integrating the PDE) that yield mismatches between a numerical solution and the actual solution of the PDE \cite{oberkampf2010verification}. 
Furthermore, these two sources of inconsistencies between the data and the underlying governing equations the PINN aims to solve might also not be aligned to each other in terms of their fidelity or accuracy. Overall, the resulting mismatches can affect the training process in non-trivial ways, decrease convergence speed, and overall degrade the performance of the trained PINN solver.\\
Although several studies have demonstrated the potential of PINNs in hybrid data–physics settings \cite{raissi2019physics}, systematic analysis of how the degree of data fidelity to the analytical formulation of the underlying physics affects the attainable precision of PINNs remains limited. In particular, it is still unclear how the quality of the training data --whether analytical, numerical, or experimental-- interacts with an exactly enforced PDE loss, and to what extent these mismatches impose an intrinsic limitation on the learning and posterior performance of the model. In this work, we focus on the fact that such data vs physics inconsistencies give rise to fundamental accuracy bounds, which we refer to as a \emph{consistency barrier}.

\medskip \noindent

\medskip \noindent
To address this gap, here we conduct a controlled study on the role of data–to-PDE consistency in PINN training and introduce a unified framework to analyze how different degrees of data fidelity impact the convergence and accuracy of PINNs. As a benchmark, we consider the one-dimensional viscous Burgers equation (a canonical nonlinear convection–diffusion PDE with accessible analytical solution) and design a set of training experiments where PINNs are trained based on a PDE loss term and a data loss term, the latter having a varying degree of fidelity.  
This setting allows us to characterize the consistency barrier and to show that, even under exact physics enforcement, the attainable accuracy of a PINN is bounded by the quality of the training data.

\noindent
The remainder of the paper is organized as follows. 
Section~\ref{sec:preliminaries} summarizes the necessary theoretical background on PINN formulation and error analysis, providing the basis to understand how data–to-physics inconsistencies affect PINN training. Section~\ref{sec:methodology} describes the benchmark problem, the data generation, and the experimental setup. Section~\ref{sec:results} presents the main results, analyzing the impact of data–to-PDE consistency on convergence and accuracy, and highlighting the emergence of the consistency barrier. Finally, Section~\ref{sec:conclusion} summarizes the findings and discusses their implications for practical PINN applications.

\section{Preliminaries}
\label{sec:preliminaries}

This section introduces the theoretical background required to understand the factors that limit the accuracy of Physics-Informed Neural Networks (PINNs). We first summarize the basic PINN framework, including notation, network representation, and the formulation of the physics-informed loss. We then show how errors in the available training data propagate through the residuals in the loss function, introducing biases that impact the attainable accuracy of the network that naturally lead to the notion of a \emph{consistency barrier}, understood as a fundamental lower bound on the loss arising from mismatches between the fidelity of the governing PDE and the data used during training.

\subsection{PINN Basic Framework}
\label{sec:pinn-framework}

For concreteness, let us start by considering a one-dimensional partial differential equation defined over a spatial domain 
$\Omega \subset \mathbb{R}$,
\begin{equation}
\frac{\partial u}{\partial t} + \mathcal{N}(u) = 0, 
\qquad {x} \in \Omega,\quad t\ge0
\end{equation}
subject to appropriate boundary conditions, where $\mathcal{N}(\cdot)$ denotes a (possibly nonlinear) spatial differential operator. In this work, we focus on steady-state solutions to simplify the analysis, noting that this choice does not affect the generality of the results. In the steady-state $\frac{\partial u}{\partial t}=0$, $u\equiv u(x)$ denotes the (spatially-varying) field solution. Within the PINN framework, the true solution $u(x)$ is approximated by a neural network 
$u_\theta(x)$, where $\theta$ labels the set of all trainable parameters (weights, biases, etc). 
A (deep) neural network is a provably universal function approximator, which, after proper training, provides a continuous representation of the solution over the whole domain.
Different from standard neural networks, the training process of a PINN is driven by the minimization over the parameter space $\theta$ of a {\it composite} loss function $\mathcal{L}(\theta)$, 
constructed as a (possibly weighted) sum of individual loss contributions:
\begin{equation}
\mathcal{L}(\theta) =
\mathcal{L}_{\mathrm{PDE}}(\theta) +
\mathcal{L}_{\mathrm{BC}}(\theta) +
\mathcal{L}_{\mathrm{D}}(\theta),
\end{equation}
where $\mathcal{L}_{\mathrm{PDE}}$ contains the residual of the differential operator acting on the approximate solution $u_\theta(x)$ at the collocation points, $\mathcal{L}_{\mathrm{BC}}$ captures the mismatch between $u_\theta(x)$ and the true values at the boundary conditions, and $\mathcal{L}_{\mathrm{D}}$ measures such mismatch between the model predictions and the training data at the measurement locations.
Finding the parameters $\theta$ that minimize $\mathcal{L}(\theta)$ yield an approximation of the solution that not only approximates well the solution at the measurement locations (and works in the interpolation regime), but is also in principle
consistent with the governing equations. This formulation provides a general and flexible framework for the solution of steady-state partial differential equations, upon which problem-specific modeling and implementation choices for the optimisation method can be built.\\
Often, the boundary conditions and the training data are grouped within the $N_{\text{data}}$ training data, and the mismatch of $u_{\theta}(x)$ and the solution $u(x)$ at these points is quantified in a mean-squared error (MSE) sense, hence:

\begin{equation}
\mathcal{L}(\theta) = \frac{1}{N_{\text{col}}}\sum_{i=1}^{N_{col}}
[\mathcal{N}(u_\theta (x_i))]^2 +
\frac{1}{N_{\text{data}}}\sum_{j=1}^{N_{\text{data}}} [u(x_j) - u_\theta (x_j)]^2,
\label{eq:exact_pinn}
\end{equation}

where $x_i$ denote the $N_{\text{col}}$ collocation points at which the PDE is enforced, while $x_j$ denote the $N_{\text{data}}$ measurement points at which training data are available or boundary conditions are imposed. Computation of the first term of $\mathcal{L}(\theta)$ at each step of any iterative optimization algorithm (in particular, all spatial derivatives appearing in the operator $\mathcal{N}(\cdot)$) benefits from automatic differentiation in the neural network.\\ 
As stated, we are dealing with what we call a {\it standard} PINN, but note that we can also build a {\it parametric} version of the problem, in which we allow the differential operator to explicitly depend on additional set of parameters $\nu$ (e.g. physical parameters). This leads to a more general PDE, whose steady-state formulation reads
\begin{equation}
    \mathcal{N}(u(x;\nu))=0,
    \qquad
    x \in \Omega, \mathbf{\nu} \in \mathcal{P},
    \label{param_PDE}
\end{equation}
where \(\mathcal{P}\) denotes a generic parameter space. In this work we will study both standard and parametric PINNs.

\subsection{Label error propagation into the loss function}
\label{sec:error-theory}
As outlined in the introduction, any set of data $\{x,u(x;\nu)\}$ used within a PINN framework might evidence different degrees of inconsistency with the ground truth given by the PDE, due to discretization-induced biases for numerical data or by measurement errors for experimental data. Formally, we denote (true) $u(x)$ the analytical solution of the PDE at $x$ and $\tilde{u}(x)$ to the solution offered by the data, such that formally

\begin{equation}
\label{eq:error}
\varepsilon({x_k}) := \tilde{u}({x_k}) - u({x_k})
\end{equation}
quantifies the data label residual inconsistency at any given measurement point $x_k$. Such inconsistencies propagate into the loss function, affecting the approximation $u_\theta(x)$ that the PINN achieves during training. Indeed, the {\it effective} loss function we get when using inconsistent data $\{x,\tilde{u}(x)\}$ is

\begin{equation}
\mathcal{L}_{\text{eff}}(\theta) = \frac{1}{N_{\text{col}}}\sum_{i=1}^{N_{\text{col}}}
[\mathcal{N}(u_\theta (x_i))]^2 +
\frac{1}{N_{\text{data}}}\sum_{j=1}^{N_{\text{data}}} [\tilde{u}(x_j) - u_\theta (x_j)]^2.
\label{eq:approx_pinn}
\end{equation}

Using Eqs. \ref{eq:exact_pinn} and \ref{eq:error}, we easily find
\begin{equation}
 \mathcal{L}_\text{eff}(\theta) = \mathcal{L}(\theta) +\frac{1}{N_{\text{data}}}\sum_{j=1}^{N_{\text{data}}} 2\varepsilon(x_j)[u(x_j)-u_{\theta}(x_j)]+ \frac{1}{N_{\text{data}}}\sum_{j=1}^{N_{\text{data}}} \varepsilon(x_j)^2,
\label{eq:error_pinn}
\end{equation}
  where $\mathcal{L}_\text{eff}(\theta)\to \mathcal{L}(\theta)$ only when all data inconsistency is removed and $\varepsilon(x_k)\to 0 \ \forall x_k$. The additional terms emerging in $\mathcal{L}_\text{eff}(\theta)$ have an effect on the optimization of $u_\theta(x)$, and thus aiming the minimize $\mathcal{L}_\text{eff}(\theta)$ instead of $\mathcal{L}(\theta)$ bias the learning process away from the true target $u(x)$ at all measurement points, and this in turn also impacts the approximation of the full-domain solution $u_\theta(x)$. Since it is in practice difficult to quantify and control the expressions of $\varepsilon(x_k)$, the resulting bias can be understood in terms of a consistency barrier.\\
  Observe that when $\nu$ is not affected by measurement errors (i.e. when using numerical data), this very same effective loss function holds for the parametric problem $u(x;\nu)$, while when using experimental data, $\nu$ itself is affected by uncertainty and additional terms emerge in the effective loss.

\section{Methodology}
\label{sec:methodology}

This section describes the experimental framework used to investigate the influence of data–PDE consistency on the performance of Physics-Informed Neural Networks (PINNs). We first introduce the benchmark problem used in this study, the viscous Burgers equation with a manufactured solution, which provides a smooth and controlled test for evaluating PINN accuracy. We then detail the generation of datasets with varying levels of fidelity and the four associated consistency scenarios where PINN are trained, as a controlled way to systematically manipulate the consistency between the PDE residual and the data term and to quantify the impact of data fidelity. Finally, we present the PINN architecture, loss formulation, and training setup used for both standard and parametric configurations. 

\subsection{Benchmark Problem}
\label{sec:test-problem}
We consider the one-dimensional viscous Burgers equation as a benchmark problem to evaluate the influence of data–to-PDE consistency on PINN's performance. This PDE is widely used as a canonical model of nonlinear transport and dissipative processes, serving both as a simplified limit of the Navier–Stokes equations and as a minimal prototype across diverse fields including turbulence and energy cascade studies, nonlinear acoustics, traffic or network flows, among others. To enable systematic analytical control of its solution and residual errors, we employ the method of manufactured solutions (MMS) \cite{sahli2020physics}, so that the PDE is augmented with a forcing source term $\mathscr{S}$ such that the resulting PDE reads
\begin{equation}
\frac{\partial u}{\partial t} + u\,\frac{\partial u}{\partial x}
= \nu\,\frac{\partial^2 u}{\partial x^2}
+ \mathscr{S}(x,\nu),
\qquad x \in [-1,1],
\label{manu_PDE}
\end{equation}
subject to Dirichlet boundary condition
\begin{align}
\label{eq:u_mms_-1}
u(-1; \nu) &= u_{\mathrm{MMS}}(-1; \nu).
\end{align}
Although the viscous Burgers equation is second order in space, only the inflow boundary condition is explicitly prescribed in this work, while the outflow boundary condition at $x=1$ is left free and closed naturally by the operator, which is sometimes known as \emph{a do-nothing} boundary condition \cite{papanastasiou1992new}. 
We will also focus in the steady state solution, i.e. $u\equiv u(x;\nu)$ denotes the solution field, $\nu$ is the viscosity parameter, and $\mathscr{S}(x,\nu)$ is the forcing term, defined such that the manufactured analytical solution $u(x;\nu)=u_{\mathrm{MMS}}(x;\nu)$, where
\begin{equation}
\label{eq:mms}
u_{\mathrm{MMS}}(x;\nu) 
= \sin(2\pi x) 
+ 0.5\,\log\!\left(\frac{1}{\nu}\right)\sin(6\pi x)
+ \log\!\left(\frac{1}{\nu}\right) + 2
\end{equation}
satisfies the governing PDE (Eq.~\ref{manu_PDE}) exactly for all $\nu$.\\
This MMS-based construction offers three main advantages. First, it yields a smooth yet nontrivial solution for any viscosity value, enabling exact evaluation of the PDE residual for a given $u_\theta(x)$ while avoiding complications associated with shocks or steep gradients.  Second, this solution also exhibits a nontrivial (nonlinear) dependence on viscosity, reflecting realistic parametric sensitivity while remaining analytically tractable. And finally, it allows to explicitly compute the discretization error of both the finite difference solver and the PINN predictions.

\medskip
\noindent 
As outlined before, in this work, we will tackle independent analysis of the standard and parametric PINNs. For the standard case, we fix the viscosity parameter to a constant value $\nu = 10^{-2}$. Fixing $\nu$ enables controlled comparisons across datasets of varying fidelity. In this case, the only input feature to the PINN is the position $x$, while the only output feature is the velocity $u_\theta(x)$. In contrast, the parametric PINN --which deals with  Eq.~\ref{param_PDE}-- is simultaneously trained on multiple viscosity values, sampled logarithmically  in the range $\nu \in [10^{-6}, 10^{-1}]$. In this case, viscosity is treated as an additional independent input feature (i.e. the input space is two-dimensional), and the learned solution $u_\theta(x; \nu)$ allows us to interpolate over space and physical conditions, enabling a richer analysis of consistency effects. In order to have features of similar orders of magnitude, the PINNs are actually fed with logMinMax-scaled viscosities instead of actual viscosities, which then range in $[-1,1]$.

\subsection{Data Generation for all consistency scenarios}
\label{sec:data-generation}

 To assess the influence of data fidelity on PINN performance, we construct three numerical datasets generated using a finite-difference solver with different degrees of fidelity (\texttt{C1-C3}), and an additional one where the data exactly matches the analytical solution, Eq.~\ref{eq:mms}, (\texttt{Analytical}, yielding a negligible numerical error and thus leading to a vanishingly small $\varepsilon(x)$)). These four datasets define our consistency scenarios that will allow us later to assess the performance of standard and parametric PINNs. Below we detail how these sets are split into training and test data and we also discuss the generation of collocation points and boundary-condition data required for PINN training.
 \\

\noindent {\bf Numerical data generation--} For datasets \texttt{C1-C3}, the corresponding solution $\tilde{u}(x;\nu)$ is obtained by numerically integrating Eq.~\ref{manu_PDE} using a stable explicit finite-difference scheme for each value of the viscosity $\nu$, logarithmically sampled in the interval $[10^{-6},10^{-1}]$. The diffusive term in Eq.~\ref{manu_PDE} is discretized using second-order central differences, while the convective term is approximated with a second-order upwind scheme. Since pseudo-time stepping is used solely as an iterative solver to reach the steady-state solution $\partial u/\partial t =0$, and since all simulations are run until convergence of the discrete residual, the final steady-state numerical solutions are independent of the specific choice of time-step size. Accordingly, any difference observed in $\tilde{u}(x;\nu)$ for datasets \texttt{C1}, \texttt{C2} and \texttt{C3} arise exclusively from the use of different spatial discretizations.\\
Specifically, dataset \texttt{C1} is generated using a coarse spatial mesh with $N^{\text{nodes}}=81$ nodes, which results in low-fidelity numerical solutions and, consequently, relatively large discretization errors $\varepsilon(x)$. Dataset \texttt{C2} employs a finer spatial mesh of $N^{\text{nodes}}=641$ nodes, yielding medium-fidelity solutions, while dataset \texttt{C3} is obtained using a highly refined spatial mesh with $N^{\text{nodes}}=7121$ nodes, leading to high-fidelity solutions and significantly reduced discretization errors. 

\begin{figure}[htb!]
    \centering
    \includegraphics[width=0.85\linewidth]{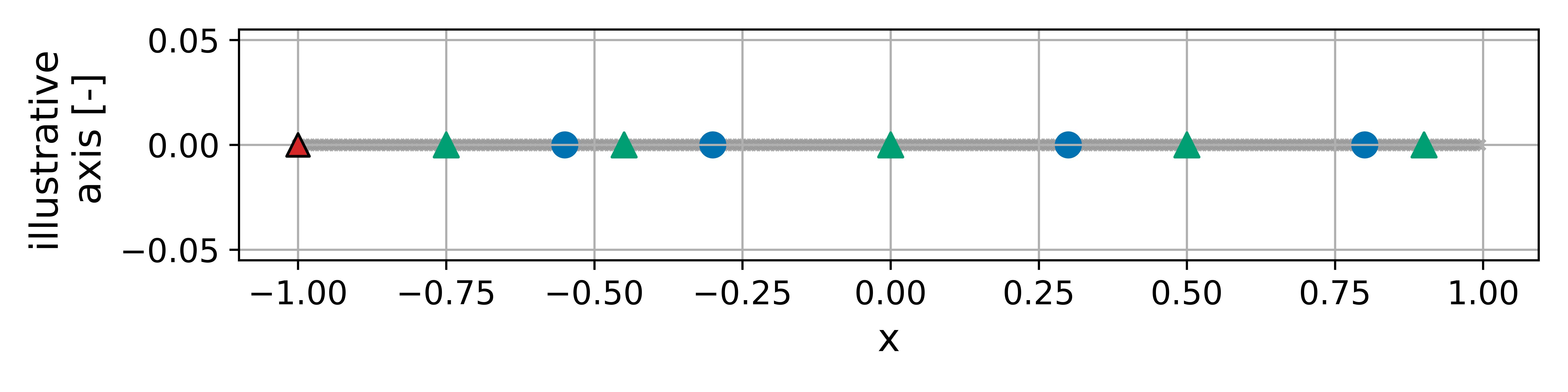}
    \includegraphics[width=0.85\linewidth]{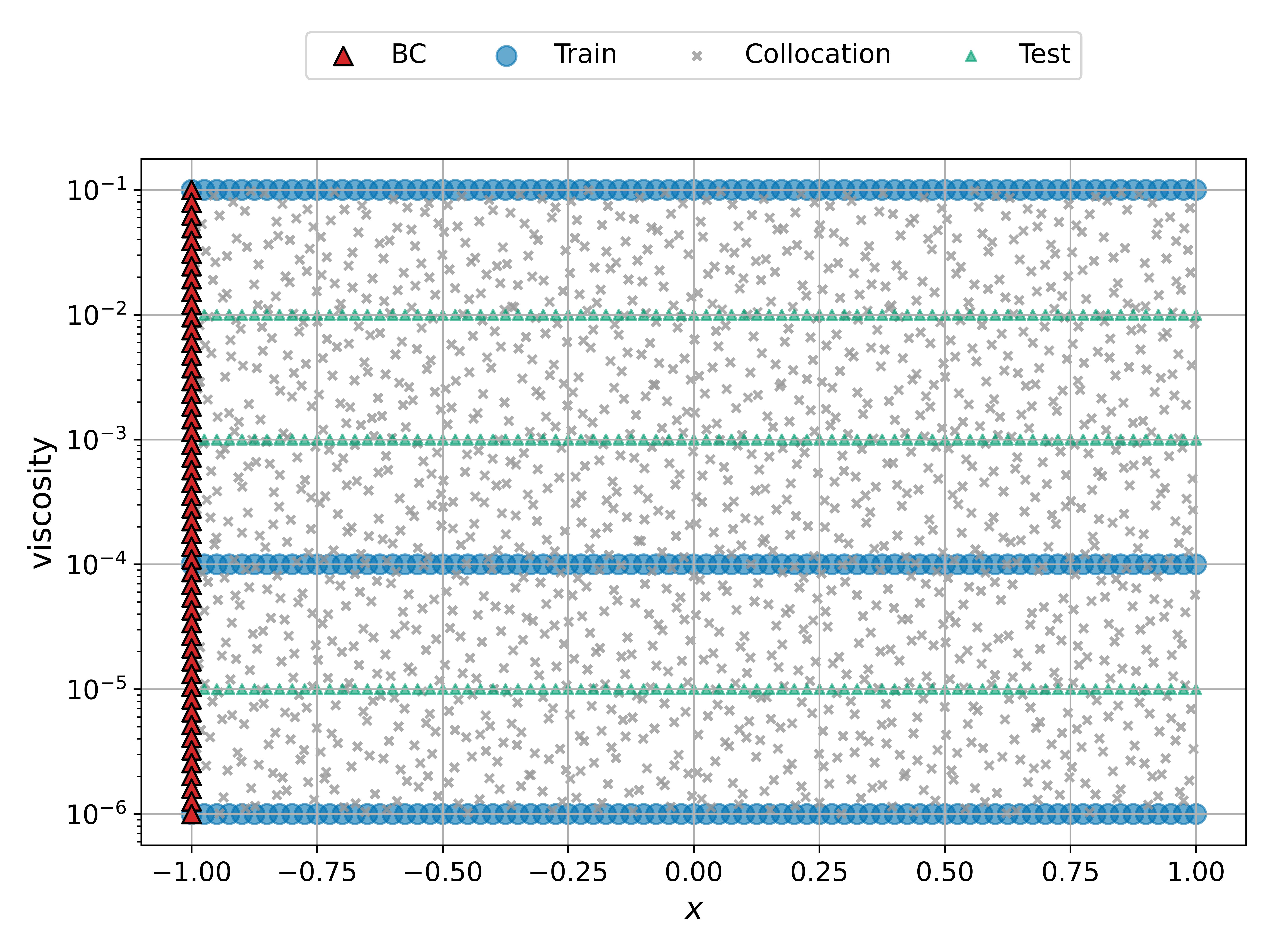}
    \caption{Distribution of all dataset points used in training and test. Top: Standard PINN. 1000 uniformed distributed collocation points, 4 Train and Test data points, and 1 BC points. Bottom: Parametric PINN, 1500 Sobol distributed collocation points, $3\times 81$ Train and Test data points, and 50 uniformly distributed BC points.}
    \label{fig:collocation}
\end{figure}

\medskip
\noindent {\bf Training data --} 
For the standard PINN (fixed $\nu$), we use $\nu = 10^{-2}$ and build a training set with 4 points with spatial location  $\{x_j\}_{j=1}^4 =\{ -0.55, -0.3,  0.3, 0.8\}$ (blue dots in the top panel of Fig.~\ref{fig:collocation}).
For the parametric PINN, we use $\nu \in \{10^{-1}, 10^{-3}, 10^{-6}\}$ (logMinMax-rescaled) and for each value of the viscosity, we define 81 spatial locations, making a total of 243 training points (blue dots in the bottom panel of Fig.~\ref{fig:collocation}). In \texttt{C1}, these 81 locations correspond to all mesh nodes used in the numerical integration, whereas for \texttt{C2} and \texttt{C3} we choose the mesh nodes which are closer in space to the \texttt{C1} mesh nodes (see Fig.~\ref{fig:collocation}). 
This controlled data generation allows us to quantify the discretization error $\varepsilon(x)$ associated to each training point, from which a mean-square error (akin to the last term in Eq.~\ref{eq:error_pinn}) is obtained, see Table~\ref{tab:data-sources}.

\medskip\noindent 
{\bf Test data and PINN double evaluation --}
For all three datasets $\texttt{C1-C3}$, the test data is conformed by numerical data that had not been used in the training. For the standard PINN (fixed $\nu$), we use again $\nu = 10^{-2}$ and build a test set with 5 points at locations  $\{x_j\}_{j=1}^4 =\{ -0.75, -0.45, 0, 0.5, 0.9\}$ (green triangles in the top panel of Fig.~\ref{fig:collocation}). For the parametric PINN, we assess the ability of the PINN to interpolate in $\nu$, and thus use $\nu \in \{10^{-2}, 10^{-4}, 10^{-5}\}$ (logMinMax-rescaled) and, just like in training data, we use 81 spatial locations for each viscosity, for a total of 243 test points (green triangles in the bottom panel of Fig.~\ref{fig:collocation}).\\
Now, how to assess the performance of each PINN once trained? We have two options: check their prediction against the solution offered in the (numerical) test data (which are also inconsistent), or against the analytical solution (fully consistent, but at odds with test data). Each evaluation would reveal different insights, and in this work we perform this double evaluation, using both pointwise global metrics (e.g. root mean-square error) and more nuanced metrics, as we will explain in the results section.

\medskip \noindent
{\bf Collocation points --} These are the spatial locations where $\mathcal{L}_\text{PDE}$ is computed. In the standard PINN configuration, 1000 equidistant collocation points are used alongside the spatial coordinate (gray crosses in the top panel of  Fig.~\ref{fig:collocation}). In the parametric PINN case, 1500 collocation points are sampled in the augmented input space using a Sobol sequence, with the viscosity dimension logarithmically rescaled (gray crosses in the bottom panel of Fig.~\ref{fig:collocation}). 

\begin{table}[htb!]
\centering
\begin{tabular}{lccc}
\toprule
Scenario & Data Source & $\frac{1}{N_{\text{training}}}\sum_j \varepsilon(x_j)^2$ & Consistency \\
\midrule
\texttt{C1} & Numerical - coarse mesh   & $ 8.4 \times 10^{-2}$ & Low \\
\texttt{C2} & Numerical - medium mesh & $ 2.8 \times 10^{-6}$ & Medium \\
\texttt{C3} & Numerical - fine mesh   & $ 1.4 \times 10^{-10}$ & High \\
\texttt{Analytical} & Analytical & $0$ & Exact\\
\bottomrule
\end{tabular}
\caption{Summary of all four consistency scenarios (\texttt{C1-C3} and the analytical one) alongside a quantification of the data consistency error for the standard PINN case. Data in \texttt{C1-C3} are generated by numerical integration of the PDE using increasingly finer computational meshes, yielding increasingly smaller discretization errors and thus smaller inconsistencies between the numerical solution and the true, analytical solution, as quantified in a mean-square sense.}
\label{tab:data-sources}
\end{table}

\medskip \noindent
{\bf Boundary condition points --} These points coincide with the domain end point at $x=-1$. In the standard PINN, this is just one point. In the parametric PINN, boundary conditions are always at $x=-1$ for different viscosities. We logarithmically sample a total of 50 points with viscosity ranging between $10^{-6}$ and $10^{-1}$, resulting in 50 viscosities uniformly sampled over the rescaled viscosity range $[-1,1]$ (red triangles in Fig.~\ref{fig:collocation}). The solution at these locations is calculated directly from the analytical solution as specified by Eq.~\ref{eq:u_mms_-1}, i.e. in our work the boundary conditions are free from inconsistencies.

\medskip Overall, this setup enables controlled degradation of data fidelity across \texttt{C1-C3}, providing a systematic framework to study how discretization-induced data-to-PDE inconsistencies propagate through the PINN loss and contribute to the emergence of a consistency barrier. A summary of the data sources, consistency scenarios and their associated discretization errors is provided in Table~\ref{tab:data-sources}.

\subsection{Details on PINN architecture and optimization}
\label{sec:pinn-formulation}
In this subsection we detail all aspects related to the training of the standard and parametric PINNs, including their architecture, the loss functions used in the optimization, and the optimizers.

\subsubsection{Network Architecture and Loss Function}
\label{sec:network-loss}

For the sake of study's simplicity, we use the same architecture and loss function configuration for both standard and parametric PINN. We employ a fully connected deep feedforward neural network consisting of seven hidden layers with 200 neurons each, and use a hyperbolic tangent as activation function. The input is one-dimensional ($x$) for the standard PINN case and two-dimensional ($x,\nu$) for the parametric case, allowing the network to interpolate across viscosities, while the output is always a one-dimensional field $u_\theta$.

\medskip \noindent
As advanced earlier, within PINN optimization each of the different loss terms can be weighted so as emphasize their importance along training. This is particularly important given the fact that, while the data losses (the boundary condition loss $\mathcal{L}_{\text{BC}}$ and the rest of training data loss $\mathcal{L}_{\text{D}}$) assess the simple mismatch between solutions (usually in a mean-square error sense), the PDE loss $\mathcal{L}_{\text{PDE}}$ assesses the residual of the behavior and properties of the solution, i.e. it considers not only the solution but its derivatives. In this sense, even small mismatches in fast spatially-oscillating solutions can induce large PDE residuals due to the disproportionately larger effects of the field spatial derivatives, as compared with simple residuals of the field itself. In a nutshell, $\mathcal{L}_{\text{PDE}}$ have intrinsically `different units' than $\mathcal{L}_{\text{BC}}$ and $\mathcal{L}_{\text{D}}$, hence the necessity of appropriately weighting the loss terms to counterbalance this.\\
A first solution would be to use a standard fixed-weights combination of losses, such that 
\begin{equation}
\label{eq:pareto_loss}
\mathcal{L(\theta)} = \alpha \, \mathcal{L}_\mathrm{PDE} + (1-\alpha) (\mathcal{L}_\mathrm{D} + \mathcal{L}_\mathrm{BC}),
\end{equation}
where $\mathcal{L}_{\text{PDE}}$ is equal to the first right-hand-side term in Eq.~\ref{eq:approx_pinn}, whereas both $\mathcal{L}_{\text{BC}}$ and $\mathcal{L}_{\text{D}}$ have the shape of the second right-hand-side term in Eq.~\ref{eq:approx_pinn}. The hyperparameter $\alpha$ can be manually tuned to explore the trade-off between physics adherence and data fidelity. In the results section -specifically in sec.~\ref{sec:pareto}- we investigate how different values of such hyperparameter $\alpha$ impact the training trajectories in the $(\mathcal{L}_{\text{PDE}}, \mathcal{L}_{\text{D}})$ plane -thereby estimating Pareto fronts-, following the methodology of Rohrhofer et al.~\cite{rohrhofer2023data}. This exploratory analysis will unveil that the optimal value of $\alpha$ is very much problem-dependent, and thus cannot be fixed a priori. Accordingly,  
from there on and in the rest of the analysis we employ a dynamical weighting strategy called \texttt{lbPINN}, a method inspired by multitask learning \cite{kendall2018multi,xiang2022self}, where the set of pre-factors $\sigma = \{\sigma_{\text{PDE}}, \sigma_{\text{BC}}, \sigma_{\text{D}}\}$ that weight each loss term is itself a set of trainable parameters, and thus this set is also dynamically updated along training. Specifically, instead of Eq.~\ref{eq:pareto_loss}, we use the following global loss function
\begin{equation}
\mathcal{L}(\theta, \sigma) = 
\frac{1}{2\sigma_{\text{PDE}}^2} \mathcal{L}_{\text{PDE}} + \frac{1}{2\sigma_{\text{BC}}^2} \mathcal{L}_{\text{BC}} + \frac{1}{2\sigma_{\text{D}}^2} \mathcal{L}_{\text{D}}
+ \sum_{i \in \{\mathrm{PDE},\mathrm{BC},\mathrm{D}\}} \log \sigma_i.
\label{eq:dynamic_loss}
\end{equation}
This self-adaptive weighting strategy allows the network to dynamically adjust the contribution of each loss component during training, facilitating robust convergence for datasets of varying fidelity and ensuring that no single term dominates the optimization process. Again, it is important to remark that the PDE loss term does not measure the mismatch between the solutions themselves, but rather the mismatch between their properties (e.g., derivatives), and therefore it does not necessarily have the same order of magnitude as the other two terms. In sec.~\ref{sec:pareto} we compare the training trajectories obtained via Eq.~\ref{eq:pareto_loss} with those found using Eq.~\ref{eq:dynamic_loss}, concluding that the latter indeed reach the Pareto front without needs to find a priori the best selection of the weights. Accordingly, the rest of the analysis will focus on training PINNs via Eq.~\ref{eq:dynamic_loss}.

\subsubsection{Optimization algorithms}\label{sec:optimization}
While multilayer perceptrons in deep supervised learning are traditionally optimized using some extension of (stochastic) gradient descent (e.g. Adam optimization), in the optimization of PINNs it is customary to switch between optimization strategies throughout training. Here, we follow such a hybrid strategy. Concretely, \textit{AdamW} is initially used for stable global convergence, followed by \textit{L-BFGS} for a smoother adjustment of the solution. This combination takes advantage of both the stability of first-order methods (\textit{AdamW}) and the precision of quasi-Newton methods precision (\textit{L-BFGS}). The number of epochs of each optimizer is detailed in Table~\ref{tab:training_epochs}. Note that the number of iterations performed within each epoch differs between optimizers. For the \textit{AdamW} stage, mini-batches of size 500 and 120 are used for the standard and parametric cases, respectively, resulting in 2 and 13 inner iterations per epoch. In contrast, L-BFGS is a
full-batch optimizer, where each iteration is an epoch. Therefore, we set the number of epochs to 3000 and 160000, respectively, and for visualization purposes in convergence curves (Figs.~\ref{fig:1d_test_error} and~\ref{fig:2d_test_error}) for the L-BFGS we plot results every 60 iterations (for the standard PINN) and every 400 iterations (for the parametric PINN). Consequently, the convergence curves presented in Sections \ref{sec:CONV1} and \ref{sec:CONV2} are reported in terms of optimization iterations rather than epochs.\\
Furthermore, prior to activating this hybrid optimization, an initial network \textit{warm-up} is performed. This process consists in, initially training the PINN for a few epochs like a basic MLP, i.e. we effectively shut down the PDE loss term by setting $1/\sigma_{\text{PDE}}^2=0$, we fix $\sigma_{\text{BC}}=\sigma_{\text{D}}=1$ and simply run \textit{AdamW} for a few epochs. For the standard PINN we applied this initial warm-up process just with the BC and Train Data. In the case of the parametric PINN however, we initially need a sufficient number of training data points to sample well all the viscosity range. Now, since we cannot use test data during training, we prevent such data leakage by augmenting the training data with fake (wrong) solutions at $\nu \in \{10^{-2}, 10^{-4}, 10^{-5}\}$ given by the solutions at $\nu \in \{10^{-1}, 10^{-3}, 10^{-6}\}$.

\begin{table}[htbp]
\centering
\begin{tabular}{lcccc}
\hline
\textbf{Case} & \textbf{Warm-up} & \textbf{AdamW} & \textbf{L-BFGS} \\
\hline
Standard PINN    & 50 & 2000 & 3000 \\
Parametric PINN  & 500 & 20000 & 160000 \\
\hline
\end{tabular}
\caption{Number of training epochs of each optimization part, for the standard and parametric PINNs. AdamW is a mini-batch optimizer, leading to 2 and 13 mini-batch updates per epoch for the standard and parametric cases, respectively. In contrast, L-BFGS operates in a full-batch regime, where each optimization step is computed using the entire training set.
 }
\label{tab:training_epochs}
\end{table}


\section{Results}
\label{sec:results}

This section presents the numerical results obtained for the standard and parametric PINN benchmark problems. The experiments are designed to systematically assess how variations in data fidelity influence the accuracy, stability, and convergence of the PINN solutions. Observed trends are interpreted in the context of the consistency barrier introduced in Section~\ref{sec:preliminaries}, which establishes a lower bound on the achievable loss when the fidelity of the data and the governing PDE are mismatched.

\subsection{Standard PINN Case}

We start our analysis with the standard PINN case, where viscosity is fixed. As a preliminary exploration, in Section~\ref{sec:pareto} we compare the training via Eqs.~\ref{eq:pareto_loss} and \ref{eq:dynamic_loss}. Interpreting the minimization of Eq.~\ref{eq:pareto_loss} as a scalarized multi-objective optimization, we construct its Pareto fronts, and show that training via Eq.~\ref{eq:dynamic_loss} reaches such front without needs to find the optimal weights a priori. Accordingly, in Sections~\ref{sec:CONV1} and \ref{sec:error1} we use the dynamically weighted loss in Eq.~\ref{eq:dynamic_loss}, and assess both the convergence curves and the performance of this PINN with respect to the different data consistency scenarios. 

\subsubsection{Optimal balance of physics vs data terms in the loss function: fixed-weight vs \texttt{lbPINN} comparison}\label{sec:pareto}
In any multi-objective optimization (MOO) problem, there exist a trade-off solution --so-called Pareto-optimal solution-- where you cannot improve one objective without making at least one of the other objectives worse. The so-called Pareto front is the set of all Pareto-optimal solutions.\\
It is suggestive to interpret PINN optimization as an MOO problem, where one objective is to minimize the PDE residual $\mathcal{L}_\text{PDE}$ and another one is to minimize the data-related loss $\mathcal{L}_\text{D}$ and $\mathcal{L}_\text{BC}$. Merging both problems via a global loss function such as Eq.~\ref{eq:pareto_loss} turns the original MOO problem into a {\it scalarized} problem. In this case, and for a fixed trade-off weight $\alpha$, we perform the optimization (following the warm-up $\to$ AdamW $\to$ L-BFGS approach discussed above), and we collect the pairs $(\mathcal{L}_\text{PDE}, \mathcal{L}_\text{D})$ reached by the algorithm throughout all optimization steps. This constitutes the set of {\it feasible} solutions, in optimization jargon. The point $(\mathcal{L}_\text{PDE}, \mathcal{L}_\text{D})$ that minimizes the global loss function in Eq.~\ref{eq:pareto_loss} for that value of $\alpha$ is a Pareto-optimal solution of the scalarized problem. Accordingly, the set of Pareto-optimal solutions found after varying $\alpha$ conforms the Pareto front.

\begin{figure}[htb!]
\centering
\includegraphics[width=1\linewidth]{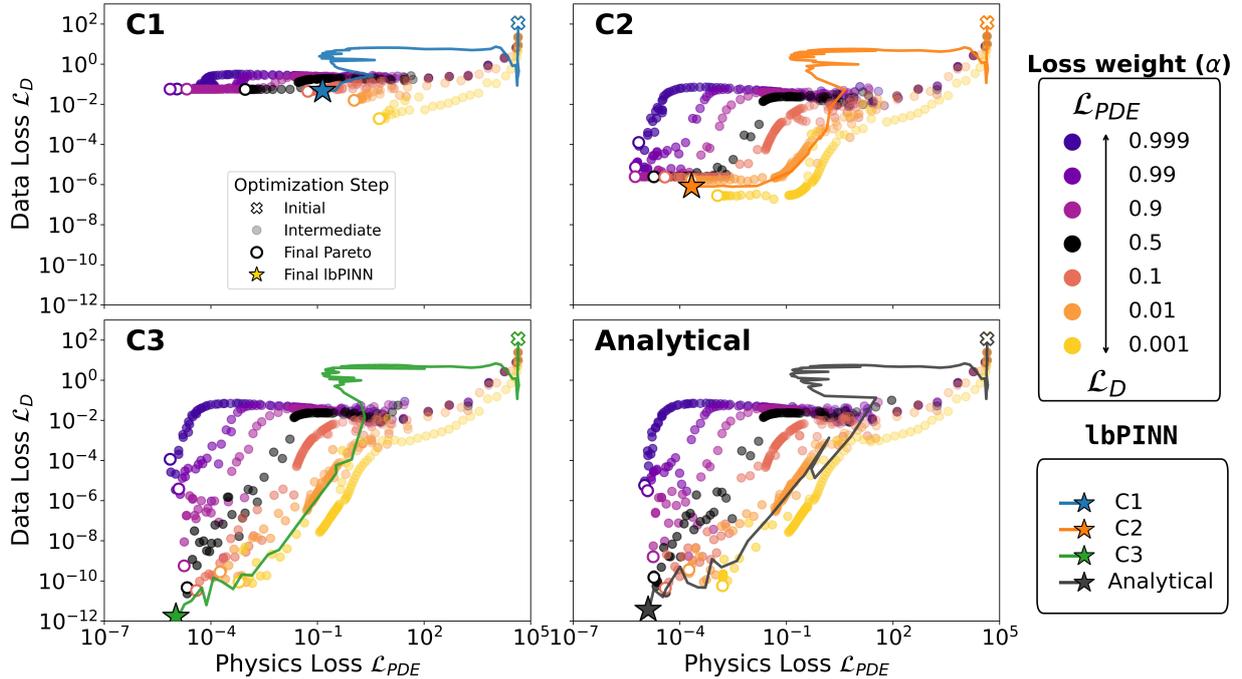}
\caption{Interpreting the optimization of the standard PINN via minimization of Eq.~\ref{eq:pareto_loss} as a (scalarized) MOO problem, we show, in the $(\mathcal{L}_\text{PDE}, \mathcal{L}_\text{D})$ plane, all feasible solutions achieved throughout optimization, for each value of $\alpha$ (color coded). The final values of each optimization (highlighted in white dots) denote Pareto-optimal solutions for each trade-off weight $\alpha$, and altogether approximate the Pareto front. Each panel depicts this information obtained for the  four consistency scenarios (\texttt{C1-C3} and \texttt{Analytical}).  Additionally, each panel also depicts the optimization trajectory of the standard PINN obtained by minimizing the Eq.~\ref{eq:dynamic_loss} given by the \texttt{lbPINN} approach (solid lines). The final solution of this optimization is indicated by a star-shaped marker in the corresponding color. Overall, \texttt{lbPINN} attains solutions that are compatible with the estimated Pareto front of the fixed-weight loss Eq.~\ref{eq:pareto_loss}.}
\label{fig:1d_pareto}
\end{figure}

\medskip \noindent 
In Figure~\ref{fig:1d_pareto} we depict, for different values of $\alpha$ (color-coded), all the feasible solutions $(\mathcal{L}_\text{PDE}, \mathcal{L}_\text{D})$ found throughout the optimization of the PINN using the loss function defined by Eq.~\ref{eq:pareto_loss}. For each $\alpha$, the Pareto-optimal solution is highlighted as a white dot, and the resulting curve approximates the Pareto-front. Each panel of this figure repeats this analysis within each of the four consistency scenarios defined previously. When $\alpha$ is decreased (and thus the PDE loss term is given relatively less importance than the data loss term), then typically optimization reaches smaller feasible values for $\mathcal{L}_\mathrm{D}$, as expected. Also, the consistency level of the training data strongly affects this behavior: higher-fidelity datasets allow to reach substantially smaller values of $\mathcal{L}_\mathrm{D}$ than low-fidelity ones. For instance, observe that $\mathcal{L}_\mathrm{D}$ can reach values of the order of $10^{-11}$ for \texttt{C3}, but only of the order of $10^{-3}$ for \texttt{C1} (and at the expense of much higher $\mathcal{L}_\mathrm{PDE}$). Intuitively, when consistency of data-to-PDE increases, the tension between minimizing $\mathcal{L}_\mathrm{PDE}$ and minimizing $\mathcal{L}_\mathrm{D}$ is reduced (both minimizations are more aligned).\\
In terms of the Pareto front, observe that in \texttt{C1} the front roughly saturates in the $\mathcal{L}_D$ axis, which is a clear indication that data is inconsistent and one cannot reduce the data fit even if augmenting its relative weight due to the clear tension between PDE and data terms, as expected. In other words, for \texttt{C1} the PDE minimization and the data loss minimization are almost incompatible. The fronts for \texttt{C2,C3, Analytical} show the classical elbow shape --which indicates compatibility between the PDE and the data--. The position of the corner of the front provides the maximum physics improvement per unit data degradation (or vice versa) and its associated value of $\alpha$ indicates a notion of trade-off optimality.
Observe also that the Pareto front for the \texttt{C3} dataset (high-resolution numerical data) and the analytical reference are very similar, certifying we don't need $\varepsilon(x)$ to be exactly zero for all $x$ for the data inconsistency to really be relevant.

\medskip \noindent 
For comparison, in the same Figure we also plot the optimization trajectories found when using the \texttt{lbPINN} loss function defined in Eq.~\ref{eq:dynamic_loss} (solid lines). Although the optimization trajectories differ from those corresponding to any fixed choice of~$\alpha$ (and this is expected as \texttt{lbPINN} explicitly allows the trade-off weights to dynamically change throughout training), the final solution reached by \texttt{lbPINN} (highlighted by a star in the figure) lies on the estimated Pareto front of Eq.~\ref{eq:pareto_loss}, or even slightly surpasses it. This finding motivates the use of \texttt{lbPINN} as the optimization strategy throughout the rest of the paper.

\subsubsection{Convergence Curves}\label{sec:CONV1}

\begin{figure}[htb!]
\centering
\includegraphics[width=0.9\linewidth]{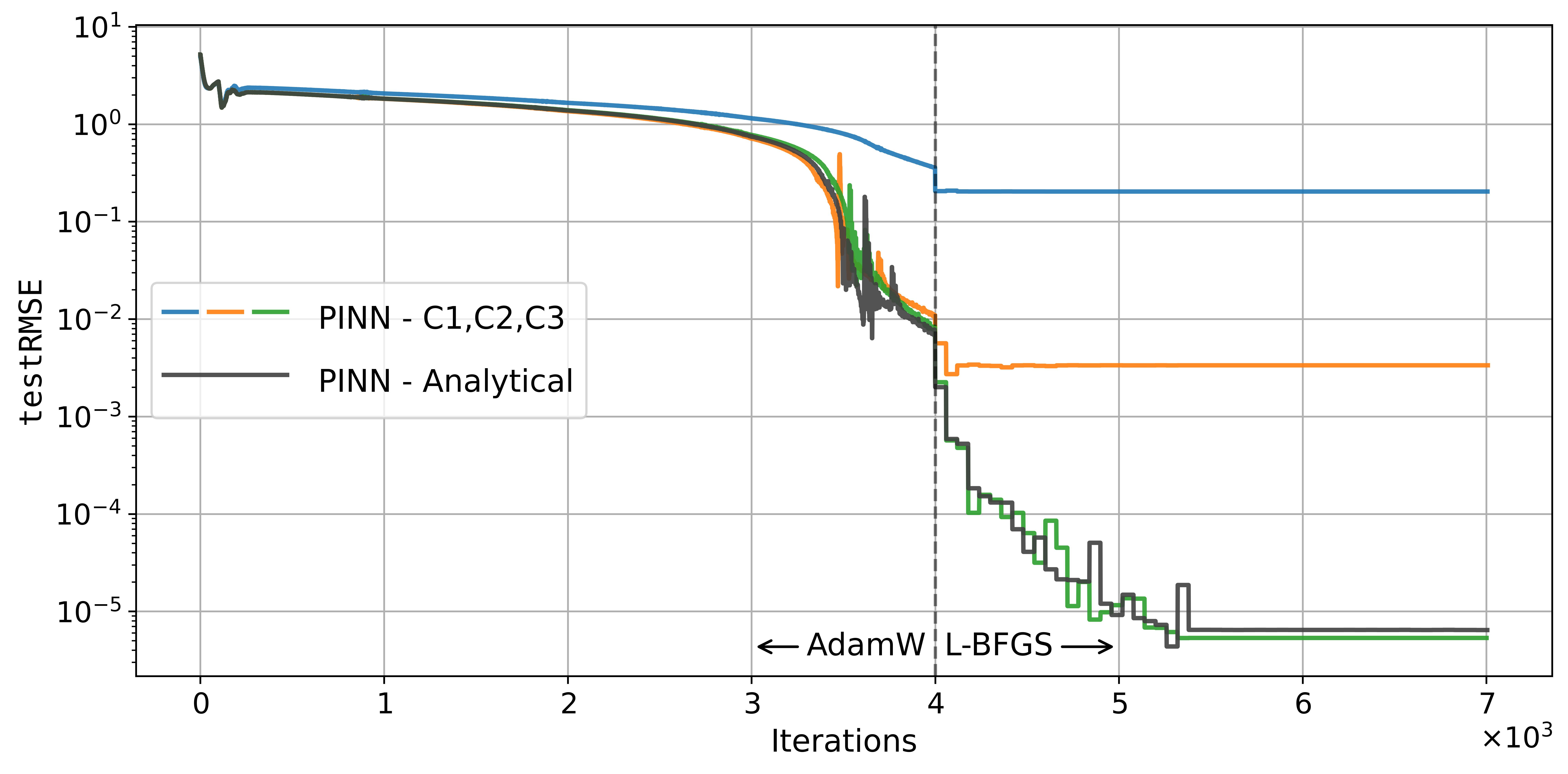}
\caption{\texttt{testRMSE} (Eq.~\ref{eq:testRMSE}, measuring the error between the PINN's prediction and the numerical data solution) as a function of the number of training iterations, for the standard PINN case trained on the four consistency scenarios. The switch between an AdamW and a L-BFGS optimizer is highlighted. }
\label{fig:1d_test_error}
\end{figure}

From now on we train the standard PINN using the dynamically weighted loss described in Section~\ref{sec:network-loss}.
Figure~\ref{fig:1d_test_error} shows, in semi-log plots, the evolution of \texttt{testRMSE} which measures the root mean square error between the PINN predictions and the numerical solutions provided the test set 
\begin{equation}
    \texttt{testRMSE} = \sqrt{\frac{1}{N_\text{test}}\sum_{k=1}^{N_\text{test}} [\tilde{u}(x_k)-u_\theta(x_k)]^2},
    \label{eq:testRMSE}
\end{equation}
as a function of the number of iterations performed within training, for all four consistency scenarios. The switch between AdamW and L-BFGS optimizers is highlighted in the figure. Observe that \texttt{testRMSE} is systematically and substantially reduced when the consistency of data is increased, and that \texttt{testRMSE} in \texttt{C3} is essentially the same as the one reached when training on analytical data, certifying that the PINN is learning and such learning converges to the actual physics when data inconsistency is reduced. Indeed, lower-fidelity datasets \texttt{C1–C2} plateau at \texttt{testRMSE} levels (note that the plot is in semi-log), whereas for \texttt{C3} and the analytical data, switching to the L-BFGS optimizer provides an extra boost in performance. 

\subsubsection{Comparing both the PINN prediction and the numerical data against analytical solution}\label{sec:error1}
\begin{figure}[htb!]
\centering
\includegraphics[width=0.9\linewidth]{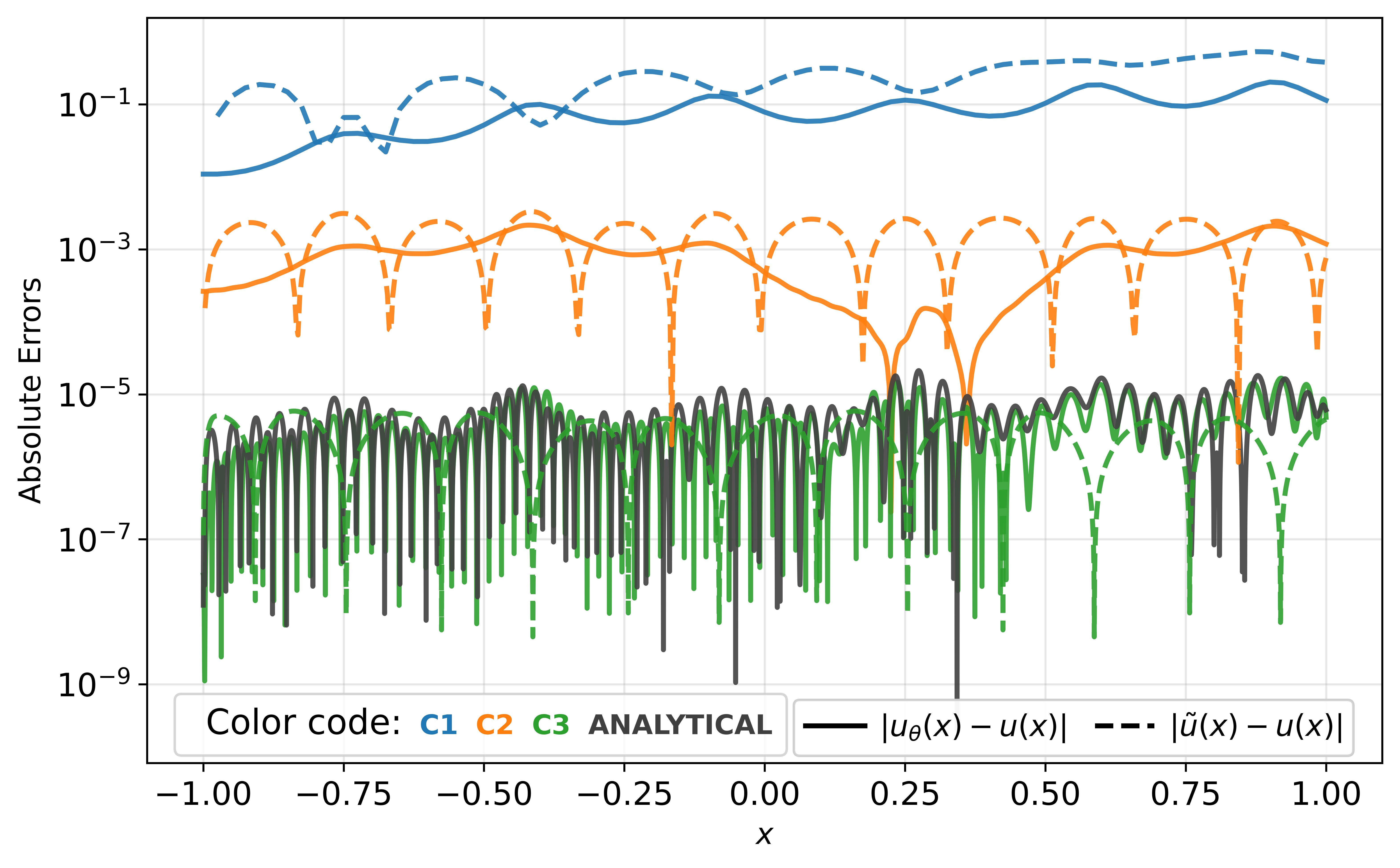}
\caption{Spatially-dependent absolute error $|u_\theta(x)-u(x)|$ (solid lines) between the PINN prediction and the analytical solution, for the standard PINN trained on the four different consistency scenarios. For comparison, in dashed lines the spatially-dependent absolute error of the numerical data $|\tilde{u}(x)-u(x)|$ is also depicted.}
\label{fig:1d_absolute_error}
\end{figure}

In the previous section we use \texttt{testRMSE} to compute how well the PINN prediction matches the (numerical) solutions in the test data, i.e. how well $u_\theta(x)$ matches $\tilde{u}(x)$, which itself shows varying degrees of inconsistencies depending on the consistency scenario under analysis. Here in turn we investigate, for each of the four consistency scenarios, how well the PINN's prediction $u_\theta(x)$ is able to deviate from $\tilde{u}(x)$ to ultimately match the true analytical solution $u(x)$. To that aim, we initially plot in Fig.~\ref{fig:1d_absolute_error} the spatially-distributed absolute error between the predicted solution  and the analytical solution $|u_\theta(x)-u(x)|$ (solid lines). The plot is in semi-log scales. PINNs trained on highly-consistent datasets (Analytical and \texttt{C3}) maintain low absolute errors throughout all domain, with minor high frequency local oscillations. In contrast, the PINN trained on lower-resolution datasets (\texttt{C1-C2}) exhibit larger deviations and larger absolute errors.

\medskip \noindent
As a complement, in the same figure we also represent (dashed curves) the spatially distributed absolute error between the numerical and the true solutions, $|\tilde{u}(x_k) - u(x_k)|$, where the locations $x_k=x_1,x_2,\dots,x_{N^{\text{nodes}}}$ correspond to the nodes of the computational meshes used in the integration scheme of each consistency scenario defined in Section~\ref{sec:data-generation}, i.e. $N^{\text{nodes}} = 81, 641, 7121$ for \texttt{C1,C2} and \texttt{C3} respectively (in the case of the \texttt{Analytical} scenario, we also use the locations of the \texttt{C3} mesh nodes). Comparing the solid and dashed lines, we conclude that the prediction of the PINN is systematically closer to the true solution $u(x)$ than the (inconsistent) numerical data $\tilde{u}(x)$ based on which the data loss term of the PINN is built. This points to the fact that the PDE loss term allows the PINN to learn the physics. At the same time, for each consistency scenario the improvement of $u_\theta(x)$ against $\tilde{u}(x)$ is not dramatic. This is highlighting the consistency barrier: while the PINN is enforcing the PDE loss that helps to discover the true $u(x)$, its ability is ultimately constrained by the (inconsistent) data loss.

\medskip \noindent
Finally, we build a global error quantifier which assesses how the prediction $u_\theta(x)$ compares with the actual analytical solution $u(x)$ by measuring the root mean square error (RMSE) between $u_\theta(x_k)$ and $u(x_k)$, where as before $x_k$ runs over the locations of the nodes of the computational meshes used to integrate the numerical solutions (in the case of the \texttt{Analytical} scenario, we use the locations of the \texttt{C3} mesh nodes). For completeness, we also report the RMSE between the numerical solution $\tilde{u}(x_k)$ and the analytical solution $u(x_k)$ in the same mesh nodes. Results are summarized in Table~\ref{tab:1d_cfd_pinn_rmse}. \\
For the lowest consistency scenario \texttt{C1} and the middle one \texttt{C2}, the PINN's global prediction error (with respect to the true solution) is only marginally smaller than the actual error of the data due to numerical inconsistencies. For high-resolution data (\texttt{C3}), the PINN's output is nearly indistinguishable from the analytical reference.

\begin{table}[htbp]
\centering
\makebox[\textwidth][c]{%
\begin{tabular}{ccc}
\toprule
Scenario &
$\sqrt{\frac{1}{N^\text{nodes}}\sum_{k=1}^{N^\text{nodes}} [\tilde{u}(x_k)-u(x_k)]^2}$ &
$\sqrt{\frac{1}{N^\text{nodes}}\sum_{k=1}^{N^\text{nodes}} [{u_\theta}(x_k)-u(x_k)]^2}$ \\
\midrule
\texttt{C1}  & 2.81e-01 & 9.65e-02 \\
\texttt{C2}  & 1.89e-03 & 1.02e-03 \\
\texttt{C3}  & 3.51e-06 & 5.34e-06 \\
\texttt{Analytical} & - & 6.61e-06 \\
\bottomrule
\end{tabular}%
}
\caption{Comparison of the RMSE (computed over the set of $N^\text{nodes}$) of (i) the numerical (inconsistent) solution against the analytical solution, and (ii) the PINN prediction against the analytical solution, for all four consistency scenarios.}
\label{tab:1d_cfd_pinn_rmse}
\end{table}

\medskip

Overall, the standard PINN analysis --including Pareto fronts, convergence curves, and error quantification-- confirms that PINN prediction accuracy is strongly dependent on training data fidelity, both when assessed against test data with the same fidelity and against analytical solutions. High-resolution datasets (\texttt{C3} and \texttt{Analytical}) consistently yield the lowest errors and most stable predictions, while lower-fidelity data result in higher errors and more pronounced spatial variations. These results establish a clear baseline for the influence of data resolution, which will guide the interpretation of subsequent parametric benchmark experiments in Section~\ref{sec:parametric_pinn}.

\subsection{Parametric PINN Case}\label{sec:parametric_pinn}

To extend the analysis of the consistency barrier beyond the Standard PINN setting, we consider a parametric PINN with two input features ($x$ and $\nu$). Using a manufactured solution, we replicate the experimental framework from the Standard PINN, training PINNs with Eq.~\ref{eq:dynamic_loss} where the data loss is built with datasets of varying fidelity (\texttt{Analytical, C1–C3}) 
while keeping the physics residual exact. This parametric setup allows us to systematically assess how the  quality of data influence model accuracy in a more complex learning task, where specially the interpolative capacity of the PINN over physical conditions ($\nu$) is more challenging.

\subsubsection{Convergence Curves} \label{sec:CONV2}
\begin{figure}[htb!]
    \centering
    \includegraphics[width=1\linewidth]{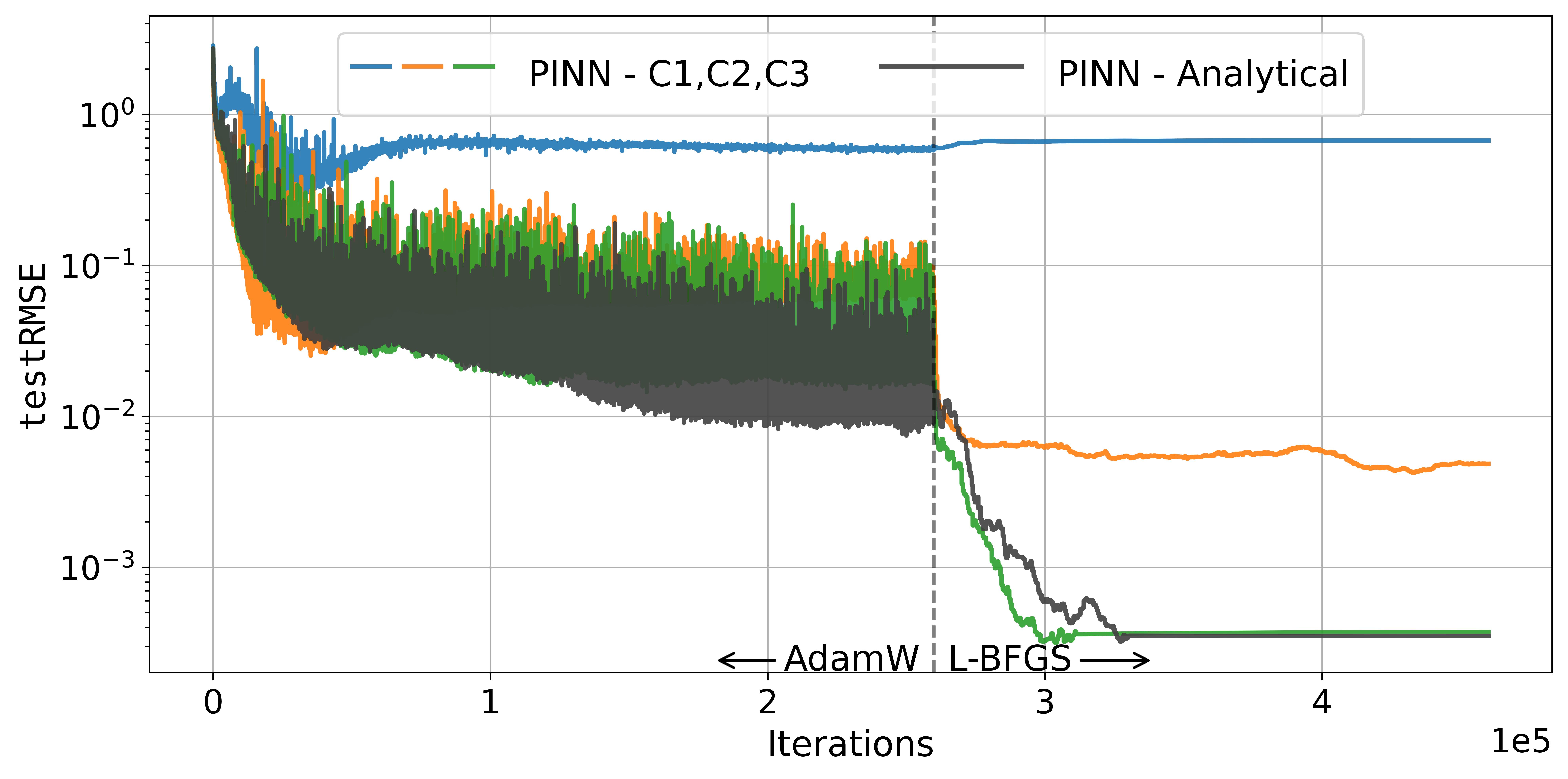}
    \caption{\texttt{testRMSE} (Eq.~\ref{eq:testRMSE}) as a function of the number of training iterations for the parametric PINN in all four consistency scenarios.}

    \label{fig:2d_test_error}
\end{figure}

Figure~\ref{fig:2d_test_error} depicts the evolution of \texttt{testRMSE} (Eq.\ref{eq:testRMSE}) measuring the mismatch between the PINN prediction $u_\theta(x;\nu)$ and the (inconsistent) test data solution $\tilde{u}(x;\nu)$ in all the points of the test set, as a function of the number of iterations of the optimization, for the parametric PINN. The behavior of the convergence curves obtained here is qualitatively similar to the standard PINN case shown in Fig.~\ref{fig:1d_test_error},
where models trained on numerical datasets with inconsistencies (\texttt{C1–C3}) saturate at progressively lower \texttt{testRMSE}. Observe that \texttt{testRMSE} saturates at slightly lower values for \texttt{C3}  than for \texttt{Analytical}: this is sensible as \texttt{testRMSE} assesses the mismatch against the (inconsistent) numerical solution in the test set, and thus such mismatch is expected to be larger for predictions $u_\theta(x;\nu)$ which are closer to the analytical solution $u(x;\nu)$. 

\medskip \noindent
The switch between optimizers is also clear, with an initial phase corresponding to AdamW followed by a smoother phase driven by L-BFGS. This second phase witnesses full convergence for cases \texttt{C1}, \texttt{C3} and \texttt{Analytical}, but observe that in \texttt{C2} the \texttt{testRMSE} curve is still mildly fluctuating even after $10^5$ iterations. A possible reason for such behavior is that, for \texttt{C2}, 
both the Data and PDE loss terms are exactly of the same magnitude order, and the tension between minimizing each of the terms is not fully dissipated, something that however might require further analysis.

\begin{figure}[htb!]
    \centering
    \includegraphics[width=1
    \linewidth]{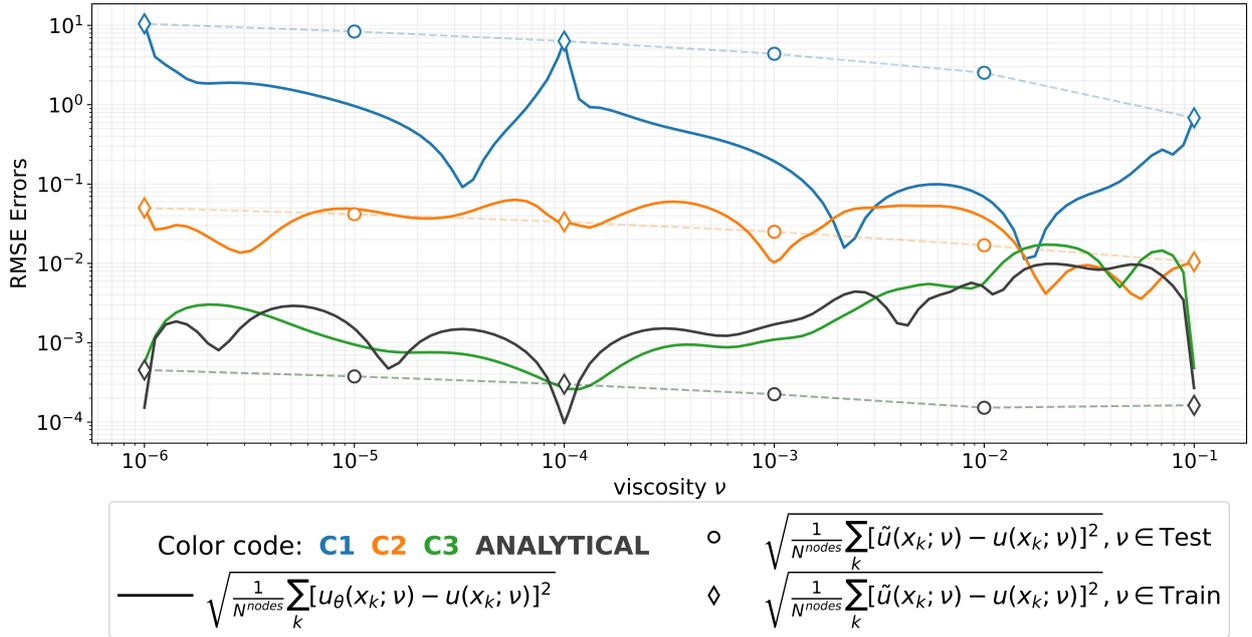}
    \caption{RMSE (over spatial locations) between parametric PINN predictions and the analytical solution (solid lines) across the range of viscosities and for the four different consistency scenarios (\texttt{C1–C3, Analytical}). Hollow points represent the RMSE (computed over all the spatial locations in the computational mesh) between the numerical solution and the analytical one, for the six viscosities $\nu \in \{10^{-6},10^{-5},10^{-4},10^{-3},10^{-2},10^{-1}\}$ (dashed lines are a guide for the eye).}
    \label{fig:2d_viscRMSE}
\end{figure}

\subsubsection{Comparing both the PINN prediction and the numerical data against analytical solution}
Like for the standard PINN, we now proceed to assess the performance of the parametric PINN not with respect to the numerical solution in the test set, but against the analytical solution. In order to gain insight on the capacity of the parametric PINN in interpolating across physical conditions, we start by considering the RMSE between the PINN prediction $u_\theta(x;\nu)$ and the analytical solution $u(x;\nu)$ (averaged over the set of spatial locations induced by the different computational meshes, as in the standard PINN case), when the parametric PINN is trained in each of the four consistency scenarios.
Fig.~\ref{fig:2d_viscRMSE} plots (in log-log scales) such RMSE as a function of $\nu$. In addition, and similar to what we did in Fig.~\ref{fig:1d_absolute_error}, in hollow dots we plot, for $\nu \in \{10^{-6},10^{-5},10^{-4},10^{-3},10^{-2},10^{-1}\}$, the RMSE between the numerical solution $\tilde{u}(x_k;\nu)$ and the true solution $u(x_k;\nu)$, computed over the same locations $x_k=x_1,x_2,\dots,x_{N^{\text{nodes}}}$ than in the previous section (dashed lines are piecewise linear interpolators, plotted for visualization purposes).\\
Comparing the solid and dashed lines for each consistency scenario reveals how the parametric PINN interpolation performance depends on data fidelity. For the low-consistency scenario \texttt{C1}, the PINN performance clearly improves upon the numerical data, while in the mid-consistency scenario \texttt{C2} the predictions are comparable to the ones obtained numerically. Highly-consistent scenarios (\texttt{C3} and \texttt{Analytical}) produce minimal differences in error, similar to the saturation observed in the standard PINN results.

\medskip \noindent
Observe also that, at high viscosities, the reduced number of collocation points --due to the logarithmic sampling-- limits the network’s ability to further reduce error, resulting in performance similar to \texttt{C2} scenario. These observations highlight a systematic relationship between training data fidelity, collocation point distribution, and achievable PINN accuracy (with respect to the true, analytical solution).

\begin{figure}[htb!]
    \centering
    \includegraphics[width=0.45\textwidth]{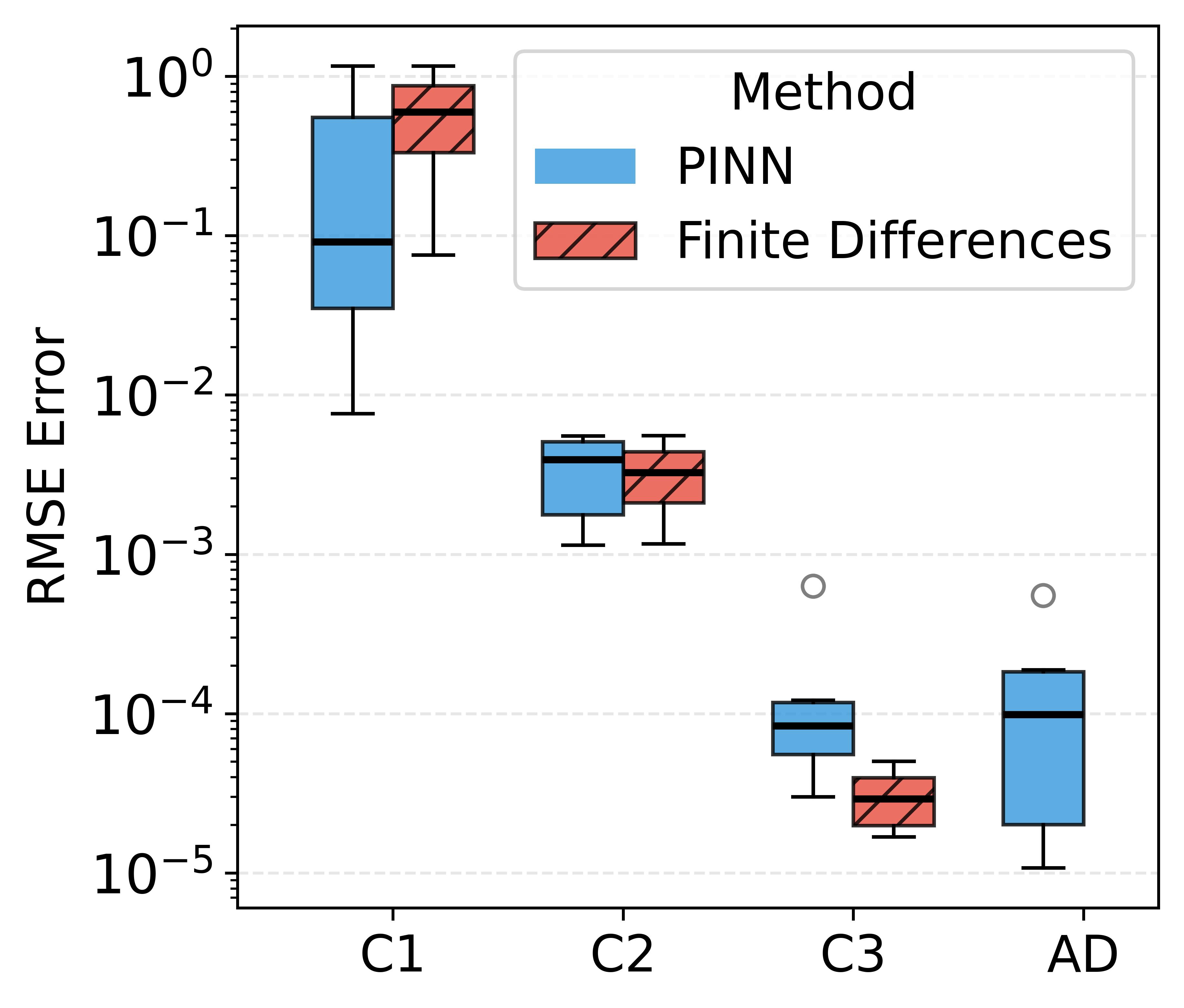} 
    \caption{(Blue) Boxplots of the viscosity-conditioned RMSE between parametric PINN predictions at the computational mesh nodes and the analytical solution, for each consistency scenario. (Red) Boxplots of the viscosity-conditioned RMSE between numerical solution at the computational mesh nodes and the analytical solution, for each consistency scenario.}
    \label{fig:2d_boxplots}
\end{figure}

\medskip \noindent
The viscosity-conditioned RMSE distribution is also shown for illustration as boxplots in Fig.~\ref{fig:2d_boxplots} (boxplots depict median, interquantile range, whiskers at $\pm 1.5$ IQR, and outliers outside whiskers). This neatly summarizes the preceding observations:
for the low-consistency scenario data \texttt{C1}, the PINN significantly learns the physics and thus deviates from the biased numerical solutions, overall reducing its error with respect to the analytical solution roughly one order of magnitude with respect to numerical. In \texttt{C2}, PINN accuracy is comparable to numerical (median test $p$-value $\approx 1$). High-fidelity scenarios (\texttt{C3} and \texttt{Analytical}) produce similar PINN accuracies, albeit showing a slightly larger error than numerical, due to the difficulty of producing systematically good prediction while interpolating across viscosities.\\

Overall, these quantitative results demonstrate that the PINN can effectively improve predictions based on low-fidelity data, while its final performance is constrained by both the fidelity of the training data and the inherent approximation limits of the network architecture. 

\section{Conclusion}\label{sec:conclusion}

Physics-informed neural networks have been proposed as a flexible framework for constructing surrogate solutions of partial differential equations by combining data with embedded physical constraints. In this work, we have shown that the effectiveness of this approach is fundamentally conditioned by the degree of consistency between the training data and the governing equations. In particular, we have identified a \emph{consistency barrier} that acts as an intrinsic lower bound on the attainable accuracy of PINN solutions when such consistency is not satisfied.\\
Using the one-dimensional viscous Burgers equation with a manufactured analytical solution as a controlled benchmark, we systematically analyzed the influence of data fidelity on PINN training. Theoretical considerations based on error propagation explain how inconsistencies in the data bias the optimization of the composite data--physics loss, leading the network toward a compromise solution rather than the exact PDE solution, even when the physics residual is enforced exactly. Our numerical experiments corroborate this observation: for low and medium-fidelity datasets, PINNs exhibit clear error saturation, with plateau levels that closely reflect the discretization error present in the data. Although the physics-based regularization enables partial correction of low-fidelity data and recovery of the dominant solution structure, the achievable improvement remains limited. In contrast, when high-fidelity numerical data or perfectly consistent analytical data are employed in training, the resulting PINN solutions converge to the analytical reference, indicating that the consistency barrier can be effectively eliminated once data errors fall below a sufficiently small threshold. Similar trends are observed for both standard and parametric PINN formulations.
Additional insight is provided by the Pareto front analysis of the data and physics losses, which reveals the inherent tension between both objectives in the presence of inconsistent data. As data fidelity increases, this tension is progressively reduced and the Pareto fronts shift toward lower attainable values of both losses.

\medskip \noindent
Overall, these findings clarify the role of data quality in physics-informed learning and explain the performance saturation often observed in practical PINN applications. From a practical perspective, our results emphasize that improving data fidelity or explicitly accounting for data errors may be as important as advances in network architectures or optimization strategies. The concept of a consistency barrier provides a useful lens for interpreting PINN results and motivates future developments aimed at mitigating the effects of data--to--PDE inconsistency. Future work should extend this analysis to higher-dimensional problems --e.g. Navier--Stokes equations--  and noisy experimental datasets, as well as accounting for the consistency of physical parameters and geometry. Another promising direction is the development of adaptive or uncertainty-aware PINN formulations that explicitly incorporate data fidelity into the loss, with the goal of mitigating the effects of the consistency barrier.\\

\noindent{\bf Acknowledgments -- } 
The authors acknowledge funding from project TIFON (PLEC2023-010251) funded by MCIN/AEI/10.13039/501100011033, Spain.
This research has received funding from the European Union (ROSAS, project number 101138319). 
This research has received funding from the European Research Council Executive Agency under TRANSDIFFUSE project, Grant Agreement No. 101167322.
Views and opinions expressed are, however, those of the authors only and do not necessarily reflect those of the European Union. Neither the European Union nor the granting authority can be held responsible for them.
LL acknowledges partial support from project CSxAI (PID2024-157526NB-I00) funded by MICIU/AEI/10.13039/501100011033/FEDER, UE,
from a Maria de Maeztu Seal of Excellence grant (CEX2021-001164-M) funded by the MICIU/AEI/10.13039/501100011033, and from the European Commission Chips Joint Undertaking project No. 101194363 (NEHIL).\\

\bibliographystyle{elsarticle-num}
\bibliography{refs}

@article{sun2020surrogate,
  title={Surrogate modeling for fluid flows based on physics-constrained deep learning without simulation data},
  author={Sun, Luning and Gao, Han and Pan, Shaowu and Wang, Jian-Xun},
  journal={Computer Methods in Applied Mechanics and Engineering},
  volume={361},
  pages={112732},
  year={2020},
  publisher={Elsevier}
}

@article{cai2021physics,
  title={{P}hysics-informed neural networks ({PINN}s) for fluid mechanics: {A} review},
  author={Cai, Shengze and Mao, Zhiping and Wang, Zhicheng and Yin, Minglang and Karniadakis, George Em},
  journal={Acta Mechanica Sinica},
  volume={37},
  number={12},
  pages={1727--1738},
  year={2021},
  publisher={Springer}
}

@article{zhao2024comprehensive,
  title={{A} comprehensive review of advances in physics-informed neural networks and their applications in complex fluid dynamics},
  author={Zhao, Chi and Zhang, Feifei and Lou, Wenqiang and Wang, Xi and Yang, Jianyong},
  journal={Physics of Fluids},
  volume={36},
  number={10},
  year={2024},
  publisher={AIP Publishing}
}

@article{eivazi2022physics,
  title={{P}hysics-informed neural networks for solving {R}eynolds-averaged {N}avier--{S}tokes equations},
  author={Eivazi, Hamidreza and Tahani, Mojtaba and Schlatter, Philipp and Vinuesa, Ricardo},
  journal={Physics of Fluids},
  volume={34},
  number={7},
  year={2022},
  publisher={AIP Publishing}
}

@article{karniadakis2021physics,
  title={{P}hysics-informed machine learning},
  author={Karniadakis, George Em and Kevrekidis, Ioannis G and Lu, Lu and Perdikaris, Paris and Wang, Sifan and Yang, Liu},
  journal={Nature Reviews Physics},
  volume={3},
  number={6},
  pages={422--440},
  year={2021},
  publisher={Nature Publishing Group UK London}
}

@article{lagaris1998artificial,
  title={{A}rtificial neural networks for solving ordinary and partial differential equations},
  author={Lagaris, Isaac E and Likas, Aristidis and Fotiadis, Dimitrios I},
  journal={IEEE transactions on neural networks},
  volume={9},
  number={5},
  pages={987--1000},
  year={1998},
  publisher={IEEE}
}

@article{raissi2019physics,
  title={{P}hysics-informed neural networks: {A} deep learning framework for solving forward and inverse problems involving nonlinear partial differential equations},
  author={Raissi, Maziar and Perdikaris, Paris and Karniadakis, George E},
  journal={Journal of Computational physics},
  volume={378},
  pages={686--707},
  year={2019},
  publisher={Elsevier}
}

@article{cai2021physics2,
  title={{P}hysics-informed neural networks for heat transfer problems},
  author={Cai, Shengze and Wang, Zhicheng and Wang, Sifan and Perdikaris, Paris and Karniadakis, George Em},
  journal={Journal of Heat Transfer},
  volume={143},
  number={6},
  pages={060801},
  year={2021},
  publisher={American Society of Mechanical Engineers}
}

@article{zhang2022analyses,
  title={{A}nalyses of internal structures and defects in materials using physics-informed neural networks},
  author={Zhang, Enrui and Dao, Ming and Karniadakis, George Em and Suresh, Subra},
  journal={Science advances},
  volume={8},
  number={7},
  pages={eabk0644},
  year={2022},
  publisher={American Association for the Advancement of Science}
}

@book{goodfellow2016deep,
  title={{D}eep {L}earning},
  author={Ian J. Goodfellow and Yoshua Bengio and Aaron Courville},
  publisher={MIT Press},
  year={2016},
  address={Cambridge, MA, USA},
  note={\url{http://www.deeplearningbook.org}}
}

@article{carleo2019machine,
  title={{M}achine learning and the physical sciences},
  author={Carleo, Giuseppe and Cirac, Ignacio and Cranmer, Kyle and Daudet, Laurent and Schuld, Maria and Tishby, Naftali and Vogt-Maranto, Leslie and Zdeborov{\'a}, Lenka},
  journal={Reviews of Modern Physics},
  volume={91},
  number={4},
  pages={045002},
  year={2019},
  publisher={APS}
}

@article{cuomo2022scientific,
  title={{S}cientific machine learning through physics--informed neural networks: {W}here we are and what’s next},
  author={Cuomo, Salvatore and Di Cola, Vincenzo Schiano and Giampaolo, Fabio and Rozza, Gianluigi and Raissi, Maziar and Piccialli, Francesco},
  journal={Journal of Scientific Computing},
  volume={92},
  number={3},
  pages={88},
  year={2022},
  publisher={Springer}
}

@article{khan2022physics,
  title={{P}hysics informed neural networks for electromagnetic analysis},
  author={Khan, Arbaaz and Lowther, David A},
  journal={IEEE Transactions on Magnetics},
  volume={58},
  number={9},
  pages={1--4},
  year={2022},
  publisher={IEEE}
}

@article{sahli2020physics,
  title={{P}hysics-informed neural networks for cardiac activation mapping},
  author={Sahli Costabal, Francisco and Yang, Yibo and Perdikaris, Paris and Hurtado, Daniel E and Kuhl, Ellen},
  journal={Frontiers in Physics},
  volume={8},
  pages={42},
  year={2020},
  publisher={Frontiers Media SA}
}

@article{huang2022applications,
  title={{A}pplications of physics-informed neural networks in power systems-a review},
  author={Huang, Bin and Wang, Jianhui},
  journal={IEEE Transactions on Power Systems},
  volume={38},
  number={1},
  pages={572--588},
  year={2022},
  publisher={IEEE}
}

@inproceedings{kendall2018multi,
  title={Multi-task learning using uncertainty to weigh losses for scene geometry and semantics},
  author={Kendall, Alex and Gal, Yarin and Cipolla, Roberto},
  booktitle={Proceedings of the IEEE conference on computer vision and pattern recognition},
  pages={7482--7491},
  year={2018}
}

@article{xiang2022self,
  title={Self-adaptive loss balanced physics-informed neural networks},
  author={Xiang, Zixue and Peng, Wei and Liu, Xu and Yao, Wen},
  journal={Neurocomputing},
  volume={496},
  pages={11--34},
  year={2022},
  publisher={Elsevier}
}

@article{rohrhofer2023data,
  title={Data vs. physics: The apparent Pareto front of physics-informed neural networks},
  author={Rohrhofer, Franz M and Posch, Stefan and G{\"o}{\ss}nitzer, Clemens and Geiger, Bernhard C},
  journal={IEEE Access},
  volume={11},
  pages={86252--86261},
  year={2023},
  publisher={IEEE}
}

@book{oberkampf2010verification,
  title={Verification and validation in scientific computing},
  author={Oberkampf, William L and Roy, Christopher J},
  year={2010},
  publisher={Cambridge university press}
}

@article{papanastasiou1992new,
  title={A new outflow boundary condition},
  author={Papanastasiou, Tasos C and Malamataris, Nikos and Ellwood, Kevin},
  journal={International journal for numerical methods in fluids},
  volume={14},
  number={5},
  pages={587--608},
  year={1992},
  publisher={Wiley Online Library}
}

@article{bonfanti2024generalization,
  title={On the generalization of pinns outside the training domain and the hyperparameters influencing it},
  author={Bonfanti, Andrea and Santana, Roberto and Ellero, Marco and Gholami, Babak},
  journal={Neural Computing and Applications},
  volume={36},
  number={36},
  pages={22677--22696},
  year={2024},
  publisher={Springer}
}

@book{ferziger2019computational,
  title={Computational methods for fluid dynamics},
  author={Ferziger, Joel H and Peri{\'c}, Milovan and Street, Robert L},
  year={2019},
  publisher={springer}
}

@article{penwarden2022multifidelity,
  title={Multifidelity modeling for physics-informed neural networks (PINNs)},
  author={Penwarden, Michael and Zhe, Shandian and Narayan, Akil and Kirby, Robert M},
  journal={Journal of Computational Physics},
  volume={451},
  pages={110844},
  year={2022},
  publisher={Elsevier}
}
\end{document}